\documentclass[runningheads]{llncs}
\usepackage{graphicx}
\usepackage{amsmath,graphicx}
\usepackage{algorithm}
\usepackage{algpseudocode}
\usepackage{caption}
\usepackage{subcaption}

\usepackage[style=ieee,maxcitenames=3]{biblatex}
\newcommand{\Figref}[1]{Fig.~\ref{#1}}
\newcommand{\Secref}[1]{Section~\ref{#1}}

\begin{document}
\title{But that's not why:
Inference adjustment by interactive prototype revision}
%
%
\author{Michael Gerstenberger\inst{1}\orcidID{0009-0008-1793-5829}
Thomas Wiegand\inst{1}\orcidID{0000-0002-1076-6777} \and
Peter Eisert\inst{1,2}\orcidID{0000-0001-8378-4805} \and Sebastian Bosse\inst{1}\orcidID{0000-0002-8104-3904}}
\authorrunning{Michael Gerstenberger et al.}
%
\institute{Heinrich Hertz Institute, 10587 Berlin, Germany \and
Humboldt University, 10099 Berlin, Germany\\
}
\maketitle              
\begin{abstract}
Prototypical part networks predict not only the class of an image but also explain why it was chosen. In some cases, however, the detected features do not relate to the depicted objects. This is especially relevant in prototypical part networks as prototypes are meant to code for high-level concepts such as semantic parts of objects. This raises the question how the inference of the networks can be improved. Here we suggest to enable the user to give hints and interactively correct the model's reasoning. It shows that even correct classifications can rely on unreasonable or spurious prototypes that result from confounding variables in a dataset. Hence, we propose simple yet effective interaction schemes for inference adjustment that enable the user to interactively revise the prototypes chosen by the model. Spurious prototypes can be removed or altered to become sensitive to object-features by the suggested mode of training. Interactive prototype revision allows machine learning na\"{i}ve users to adjust the logic of reasoning and change the way prototypical part networks make a decision.

\end{abstract}

\section{Introduction}
\label{sec:intro}
Human learning typically involves communication between individuals based on a shared understanding of symbolic mental representations \cite{Planer2021Symbolic}. However, neural networks are usually black box models. Consequently, the sub-symbolic processing in artificial neural networks has hindered analogous interactions between humans and machines. \\
 A common strategy in interactive machine learning (IML) has therefore been additional training with modified data \parencite{ho_deep_2020,wang_deepigeos_2018},  e.g.~by reusing samples with manually corrected annotations to refine  an initial prediction \parencite{ho_deep_2020}. Alternatively, a second network can be trained to predict the user feedback \parencite{wang_deepigeos_2018}\parencite{liao_iteratively-refined_2020}. In both cases the concepts learned by the network remain hidden inside the black box. Only recent developments in explainable AI allow for a better understanding of the models reasoning \cite{chen_this_2019}. This enables the interaction on the conceptual level.
 In this paper, we build on recent advances in prototype-based learning that explains predictions using previously learned prototypes \parencite{chen_this_2019}.\\
Prototype theory roots in cognitive science and suggests that human cognition relies on prototypes that structure conceptual spaces \parencite{geeraerts_prototype_2008,balkenius_spaces_2016}. In such spaces similar concepts lie closer while there is a greater distance between dissimilar ones. Evidence suggests that such an order with strong association between similar features even exists in the human brain \cite{bosking_orientation_1997}. Leveraging this similarity of natural and artificial cognitive representation, we propose a novel approach to IML that allows for direct human-AI interaction on the conceptual level.\\
Our major contribution is an interactive procedure that allows to eliminate the effect of spurious features that do not relate to the objects classified by ProtoPNet including a quantitative evaluation. We apply and compare our approach to a simple removal of prototypes and illustrate the effect it has on the latent distribution. It shows that object-sensitivity can be achieved with few clicks abolishing the need to mask training images. While concurrent related work evaluated the possibility to remove artificially introduced artifacts we show quantitatively that object-sensitivity of virtually all prototypes can be achieved for a dataset where irrelevant features correlate naturally with the class concepts \cite{bontempelli2022concept}. Moreover our approach yields good results even with fixed encoder weights reducing the computational cost and hence improving training speed during the interaction runs: It shows that inference of prototype networks can be adjusted solely by the interaction with concepts in latent space. In \Secref{sec:protopnet} we review the concept of prototype-based learning. \Secref{sec:method} introduces our interaction scheme and shows how it can be used to adjust the inference. \Secref{sec:results} presents experimental results showing how inference adjustment can be achieved. 
\begin{figure*}[t]
    \includegraphics[width=\textwidth]{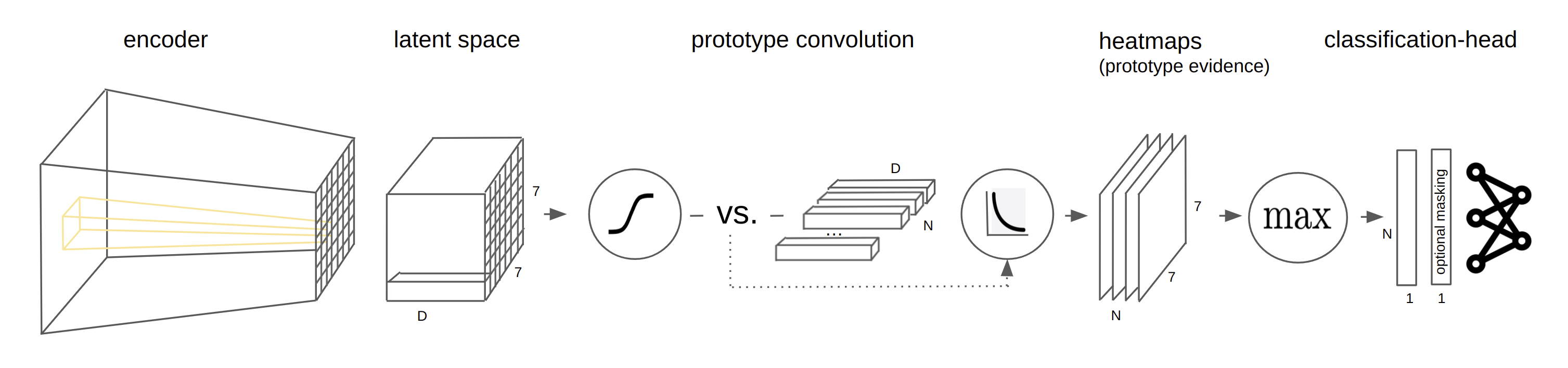}
    \caption{A prototypical-part network: Suitable prototype vectors are learned during end-to-end training. The prediction relies on the detection of prototypes in feature space. This is achieved by prototype convolution that computes the inverted distance between the grid vectors and the prototypes.}
    \label{network}
\end{figure*}

\section{Prototype-based Learning}
 \label{sec:protopnet}
Prototype-based deep learning builds on the insight that similar features form compact clusters in latent space. Prototypes are understood as concepts that are central for a class \parencite{bau2020understanding}.\\
In prototype based deep learning latent features are either represented by a single vector or a latent grid \parencite{yang_robust_2018,chen_this_2019}. In prototypical part networks a prototype refers to (1) the prototype-vector and (2) the image patch of the image with the closest grid vectors in embedding space. \Figref{network} shows the structure of the model of type ProtoPNet that was used here. Prototypes cluster feature space analogously to centroids in $k$-means. Prototype convolution computes the Euclidean distance between prototypes and feature-vectors. An evidence score is computed for each latent grid position (7x7) and prototype (N=2000) as a function of the inverted distance. In prototype networks with Gaussian Prototype Layer the eucledean distance is weighted by the covariance matrix instead \cite{gerstenberger2023differentiable}. The maximum taken over each heatmap constitutes the input to the classification head. Prototypes are pushed to the closest feature vectors such that the latent encoding of an image patch coincides with the prototype-vector and corresponds to image patches from the training set \parencite{chen_this_2019}. Hence, they can be conceptually equated with image patches from the training set. \\
There is a growing body of work on prototype based deep learning. Typically, prototype based learning is used to classify images and explain what the relevant features are. Differences exist regarding the question whether or not negative reasoning is allowed or the way prototypes are fitted. Training can be end-to-end \parencite{chen_this_2019,li_deep_2018,yang_robust_2018} or iterative \parencite{chong_towards_2021,dong_few-shot_2018}. Recent work also includes models that use configurations of prototypical parts for deformable prototypes \cite{donnelly2022deformable} or hierarchical prototypes that follow predefined taxonomies and show how prototype based decision trees can be constructed \parencite{hase_interpretable_2019}\parencite{nauta_neural_2021}. Besides classification, prototype-based deep learning is used for explainable segmentation \cite{gerstenberger2023differentiable} or tracking of objects in videos, where prototypes are retrieved from previous frames and used as sparse memory-encodings for spatial attention \parencite{ke_prototypical_2021}. Some approaches address data scarcity and have particularly been developed for few-shot problems making predictions for a new class with few labels ($n<20$). Prototypes can be used for classification or semantic segmentation \parencite{snell_prototypical_2017,dong_few-shot_2018}. The benefits of deep prototype learning are the high accuracy and robustness its potentials for outlier detection \parencite{yang_robust_2018}, the ability to yield good predictions in few shot problems \parencite{snell_prototypical_2017,dong_few-shot_2018} and an increased interpretability that allows for intuitive interaction with the represented concepts \parencite{chen_this_2019,hase_interpretable_2019}.

\section{Interactive Prototype Revision}
\label{sec:method}

\begin{figure}[htb]
  \includegraphics[width=12cm]{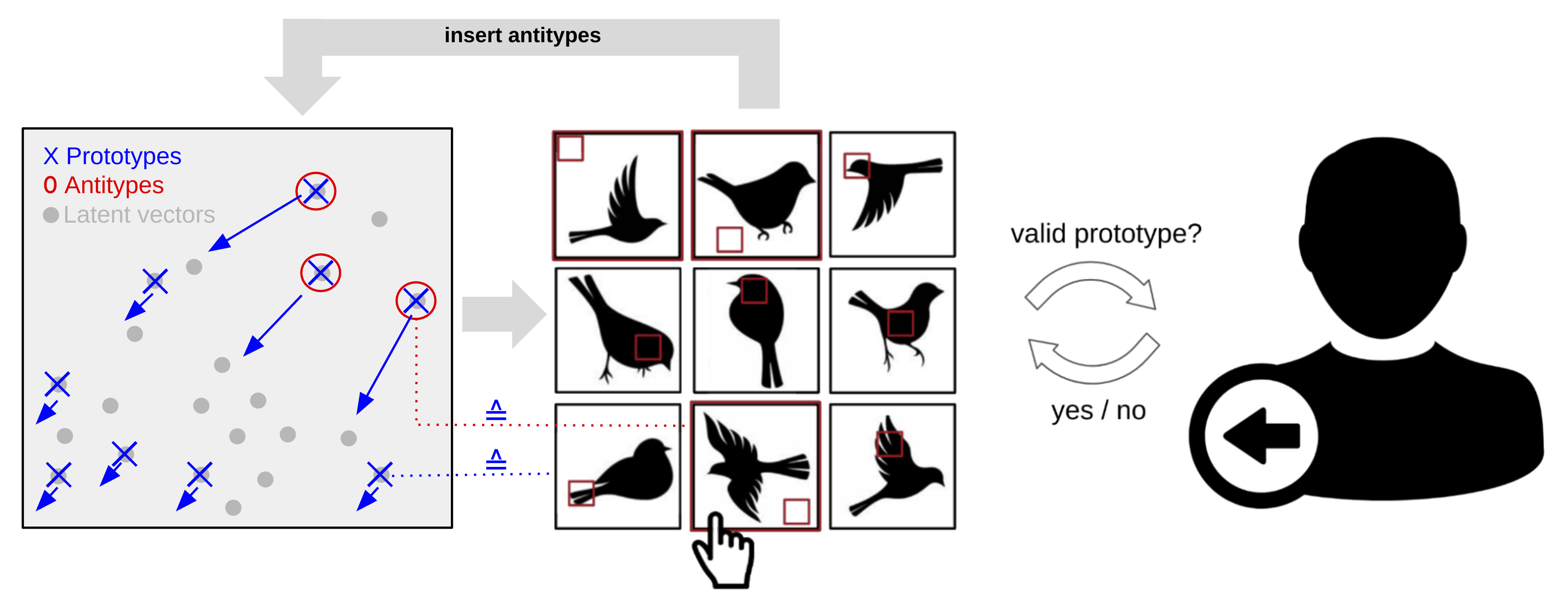}
\caption{Interactive prototype revision:
The prediction of a class is based on the detection of its previously learned prototypes. Inference is adjusted as the user identifies antitypes that exert a repelling force on prototypes.}
\label{user_interaction}
\end{figure}

We aim to adjust the inference of the model by having the user revise the prototypes. Here we compare two approaches. Inference can be adjusted (1) by masking the model's evidence for undesired prototypes or (2) by iterative prototype revision via repetitive user feedback and a custom loss-function: Prototypes are removed from areas of latent space that encode non-object patches as the user identifies antitypes i.e. vectors that code for features unrelated to the concepts represented by the prototypes (here: birds; \Figref{user_interaction}).\\
 Hence, the network acts like a student that consults the teacher to retrieve information about the learned prototypes. The user takes on the role of the teacher and either accepts or rejects the prototypes presented by the student (\Figref{user_interaction}).  \\
For deselection without replacement, a mask is assembled to exclude prototypes from inference. To adjust for a decrease in accuracy, the final layer of the network is trained afterwards. Prototype revision with replacement can be achieved with a customized loss term $\textit{Refine}$ which relies on the $L^2$ norm of the prototypes $p_j$ and the user-injected antitypes $q_s$.

The cost term measures the distance between $p_j$ and $q_s$ and is maximal if antitype and a prototype coincide. It thus encourages the prototypes to diverge from latent encodings of non-object patches.
\begin{equation*}
\operatorname{Refine} =\max_{i}\biggl[\log\left(\frac{d_i + 1}{d_i + \epsilon}\right)\biggr]\times\frac{1}{-\log{(\epsilon})}, \text{ where } d_i= \left\|\mathbf{p}_{j}-\mathbf{q_s}\right\|_{2}^{2}
\end{equation*}
\\

The antitype vectors are initialized with the prototypes that the user identifies as being of type non-object. The prototypes diverge from the antitypes that are inserted by the user due to the repelling term in the loss. \Figref{user_interaction} illustrates this process.\\ A second term $\textit{Con}$ imposes a soft-constraint and ensures that prototypes stay in the hyper-cube that contains the latent vectors by penalizing any value $\textit{i}$ of each prototype-vector $\textit{j}$ that lies outside of the desired range.

{\footnotesize
\begin{equation*}
    \operatorname{Con} = \sum_j\sum_i v_{i,j}
    \text{ where }
    v_{i,j}=
    \begin{cases}
      1, & \text{if}\ (p_{i,j} > 1) \lor (p_{i,j} < 0) \\
      0, & \text{otherwise}
    \end{cases}
\end{equation*}
}

These terms are part of the final loss. It additionally contains the cross-entropy $\textit{CrsEnt}$, the clustering cost $\textit{Clst}$ and the separation cost $\textit{Sep}$ as well as the $\textit{L1}$ term from the original loss of ProtoPNet.

 {\footnotesize
\begin{equation*}
L = \text{CrsEnt} + \lambda_{1} \text { Clst }+\lambda_{2} \text { Sep } +\lambda_{2} \text { L1 } +\lambda_{3} \text { Refine } + \lambda_{4} \text { Con }
\end{equation*}
}
 Although prototypes diverge from the deselected feature vectors, some may move towards areas in latent space that encode other spurious features. Hence repeated training for several iterations is needed to remove these prototypes. In total around 350 clicks are necessary until convergence is achieved using the CUB dataset with 200 classes (\Figref{main_results}; lower left). This is considerably less effort than the  na\"{i}ve approach of attributing pixelwise annotations to the 6200 training images, masking out irrelevant areas and training a standard model.
The user repeatedly interacts with the model and successively explores areas in latent space that are covered by non-object patches. At the beginning of each repetition the prototypes are pushed to the closest feature vector to identify the prototypical image patches. In the next step the user is consulted to reject unreasonable prototypes.\\ 
\noindent\makebox[\textwidth][c]{
\begin{minipage}{10cm}
    \begin{algorithm}[H]
	\begin{algorithmic}[1]
	    \State $ Q\leftarrow\ \emptyset $
		\For {$repetition=1,2,\ldots N $}
		    \State $ p_j\leftarrow \arg \min_{z\in Z_j}\left \| z -p_j \right \|$ \Comment{Push prototypes}
		    \State $ Q\leftarrow\ Q \cup \textit{non-object}(P)$ \Comment{Consult user}
			\For {$epoch=1,2,\ldots,M$}
			    \For {\textit{batch} of \textit{Dataset}}
			        \State  $\textit{loss} \leftarrow\ \textit{L}(y,\textit{net}(\textit{batch}),P,Q)$
			        \State $\textit{net} \leftarrow \textit{SGD}(\textit{net},\textit{loss})$  \Comment{Move prototypes}
			   \EndFor
			\EndFor
		\EndFor
	\end{algorithmic}
	\centerline{\textbf{Alg. 1:} \text{Iterative prototype revision}}
\end{algorithm}
\end{minipage}}
\\
The set of antitypes $\textit{Q}$ is united with the subset of the prototypes $\textit{P}$ that are identified as non-object prototypes and the network is trained using our deselection loss. Hence, the areas of latent space that encode spurious features are successively clustered by the interplay of user and model (Alg.~1).

\section{Results}
\label{sec:results}
\begin{figure}[ht]
    \begin{minipage}{0.6\linewidth}
        \begin{subfigure}[b]{\textwidth}
            \includegraphics[width=\textwidth,keepaspectratio]{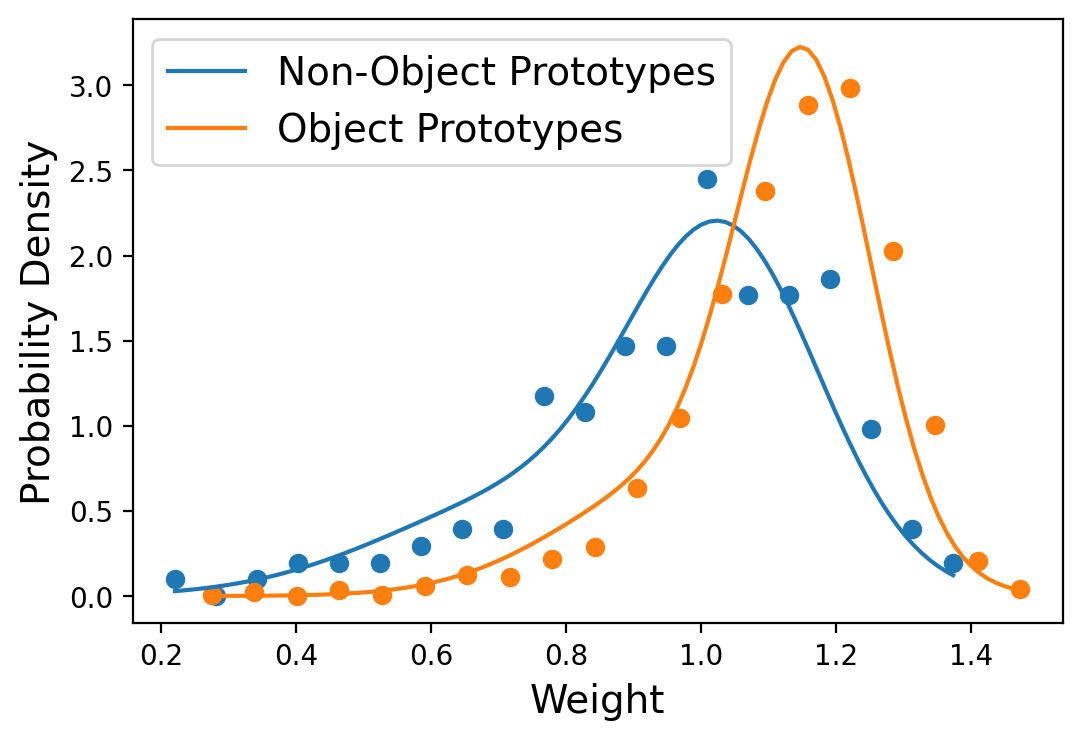}
            \caption{}
            \label{subfig:weights}
        \end{subfigure}
        \begin{subfigure}[b]{\textwidth}
            \includegraphics[width=\textwidth,keepaspectratio]{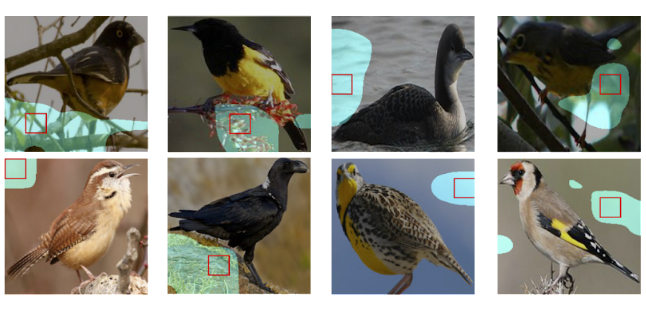}
            \caption{}
            \label{subfig:nonObjectPrototypes}
        \end{subfigure}
    \end{minipage}
    \hfill
    \begin{minipage}{0.38\linewidth}
        \begin{subfigure}[b]{\textwidth}
            \includegraphics[height=9.2cm,width=\linewidth]{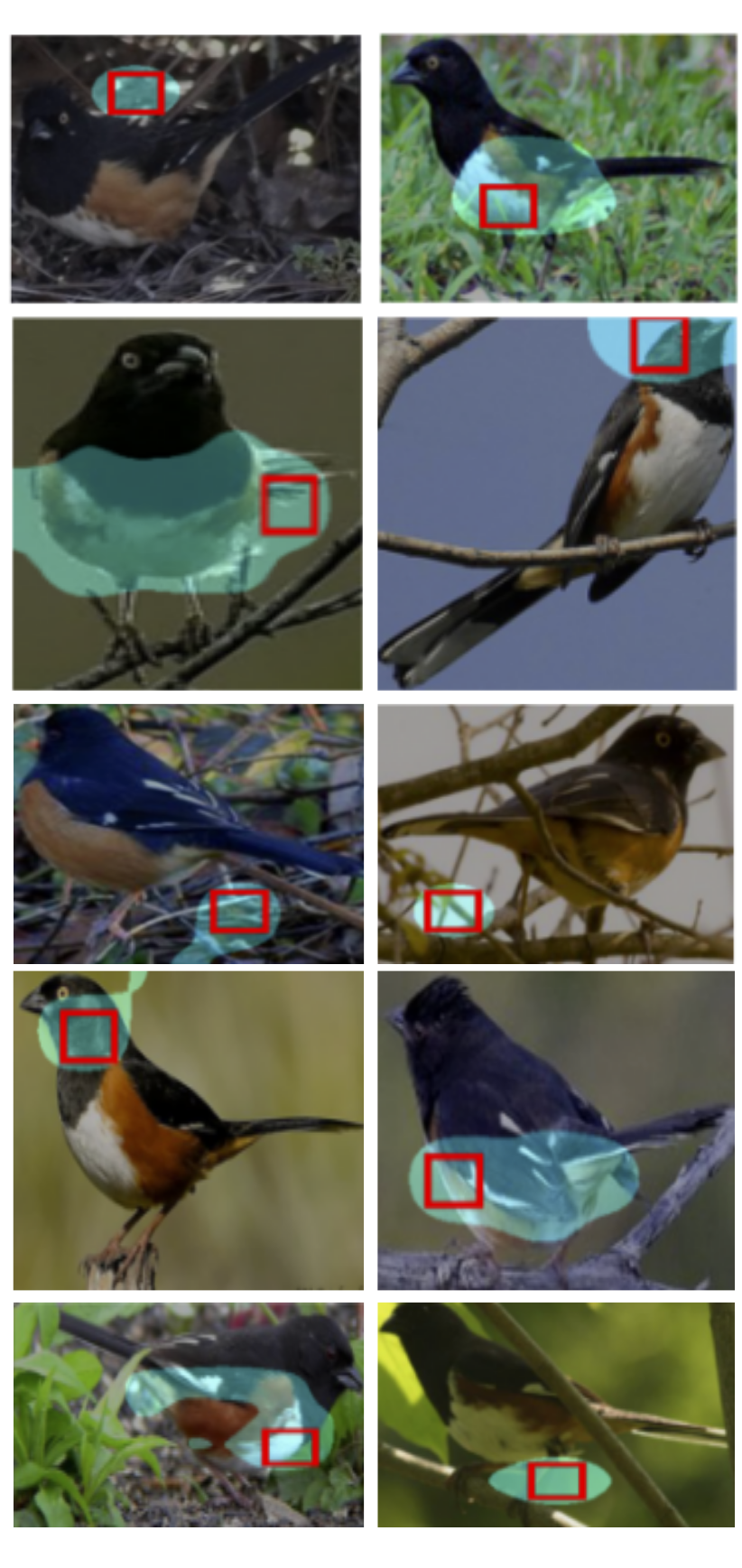}
            \caption{}
            \label{subfig:objectPrototypes}
        \end{subfigure}
  \end{minipage}
\caption{The relevance of non-object prototypes:
(a) The distribution of weights of object-prototypes and non-object prototypes in the final layer. (b) Examples for non-object prototypes (c) Prototypes of the class "Towhee".}
\label{relevance}
\end{figure}

Our experiment includes three conditions. For the baseline condition we complete the first two stages of the original training schedule of ProtoPNet \parencite{chen_this_2019}. For that purpose we use the CUB200-2011 dataset. The model is trained for five epochs with the encoder weights fixed, then all layers and the prototype-vectors are trained together. Afterwards, we push the prototypes to the closest image patch of any training image of the prototype's class. Finally the classification head is trained to adjust for the changes and allow for the L1 term to converge \parencite{chen_this_2019}.\\
We aim at a quantitative internal evaluation by dissecting different parts of the training schedule. Our experiment aims at comparing (1) the baseline condition, to (2) the presented interactive revision procedure (Alg.~1) and (3) the effect of a subsequent removal of prototypes without replacement by prototype masking. To enable reproducibility we simulate the user interaction computationally for named experiments. To this end we rely on pixelwise annotations included in the dataset and identify non-object prototypes by computing the image patch using the upscaling-rule \parencite{wah_caltech-ucsd_2011}. If there is any overlap between the patch and with the object it is considered to be an object-prototype. In contrast denote prototypes have no overlap with the depicted objects non-object prototypes.\\
 In the baseline condition between $157$ and $251$ of the $2000$ prototypes were identified as non-object prototypes  ($\bar{n}_{\textit{bg}}=206 , \bar{n}_{\textit{fg}}=1794$). \Figref{subfig:nonObjectPrototypes} shows a random selection of non-object prototypes for a sample run. Here, an overlap of at least 75\% exists only for 1077 prototypes. Analogously, \Figref{subfig:objectPrototypes} shows all prototypes for the class "Eastern Towhee". Red boxes indicate the location of the prototype. Blue areas indicate where the heatmap of prototype evidence is larger then 1\% of its maximum.

 \Figref{relevance}a shows the distribution of the weights in the final layer for object and non-object prototypes in a representative run with a Gaussian fit for the PDF. Larger weights are more frequent for object prototypes with a peak in the PDF at $w = 1.2$. However, substantial weights exist also for non-object prototypes highlighting the relevance of the respective non-object prototypes.\\

\begin{figure}[htb]
  \begin{minipage}[b]{1.0\linewidth}
\centering
\centerline{\includegraphics[width=11cm]{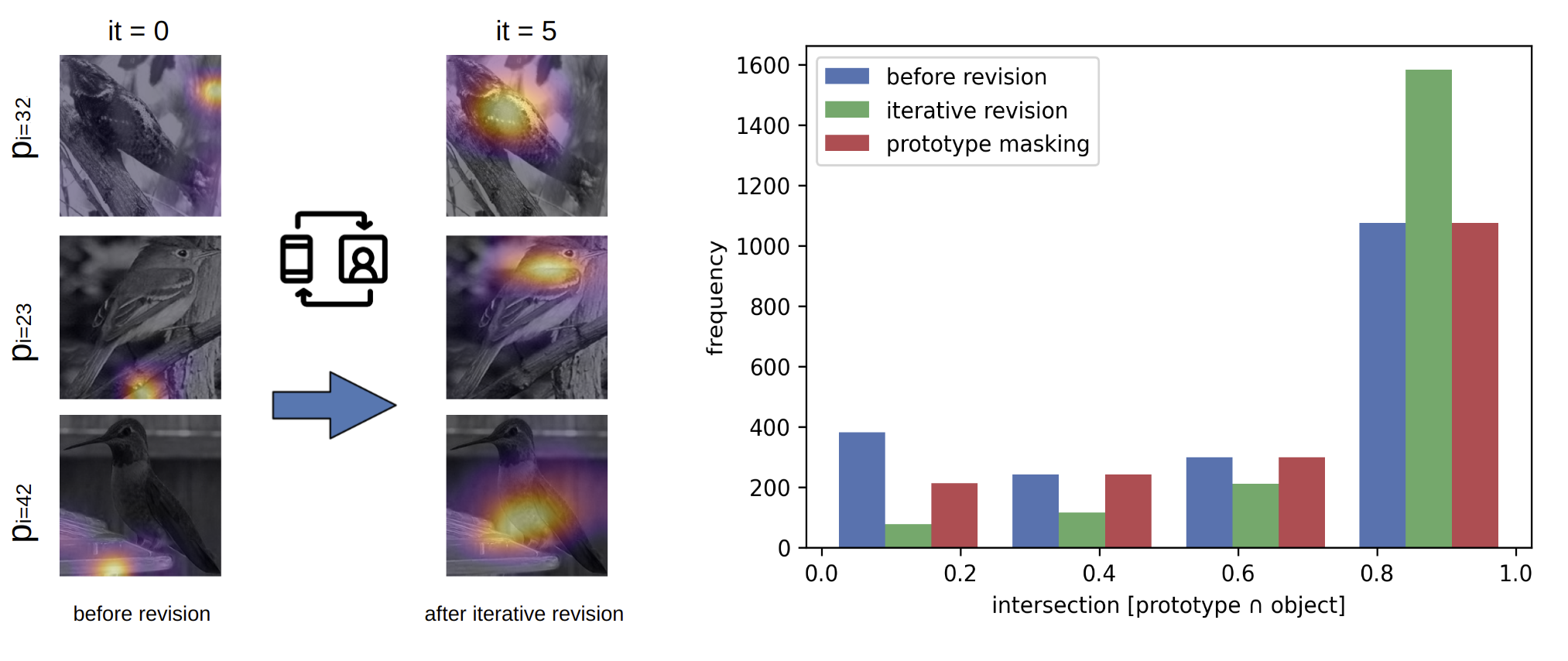}}
\end{minipage}
\begin{minipage}[b]{1.0\linewidth}
\centering
\centerline{\includegraphics[width=11cm]{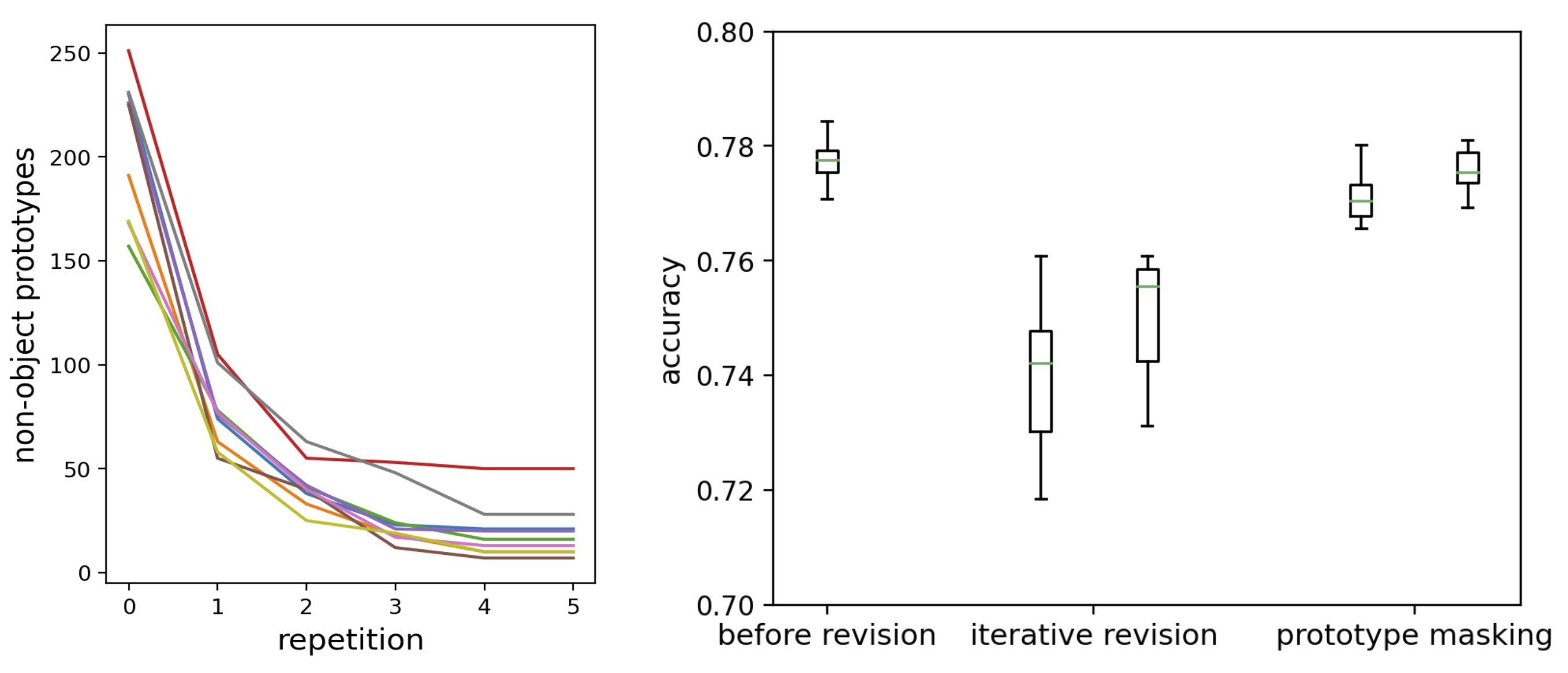}}
\end{minipage}
\caption{Prototype revision: (a) Non-object prototypes become sensitive to object features upon iterative revision (upper row). The number of non-object prototypes during iterative revision (lower left). The change in accuracy for different schemes before and after fine-tuning of the last layer (lower right).}
\label{main_results}
\end{figure}
\begin{figure}[htb]
  \includegraphics[width=12cm]{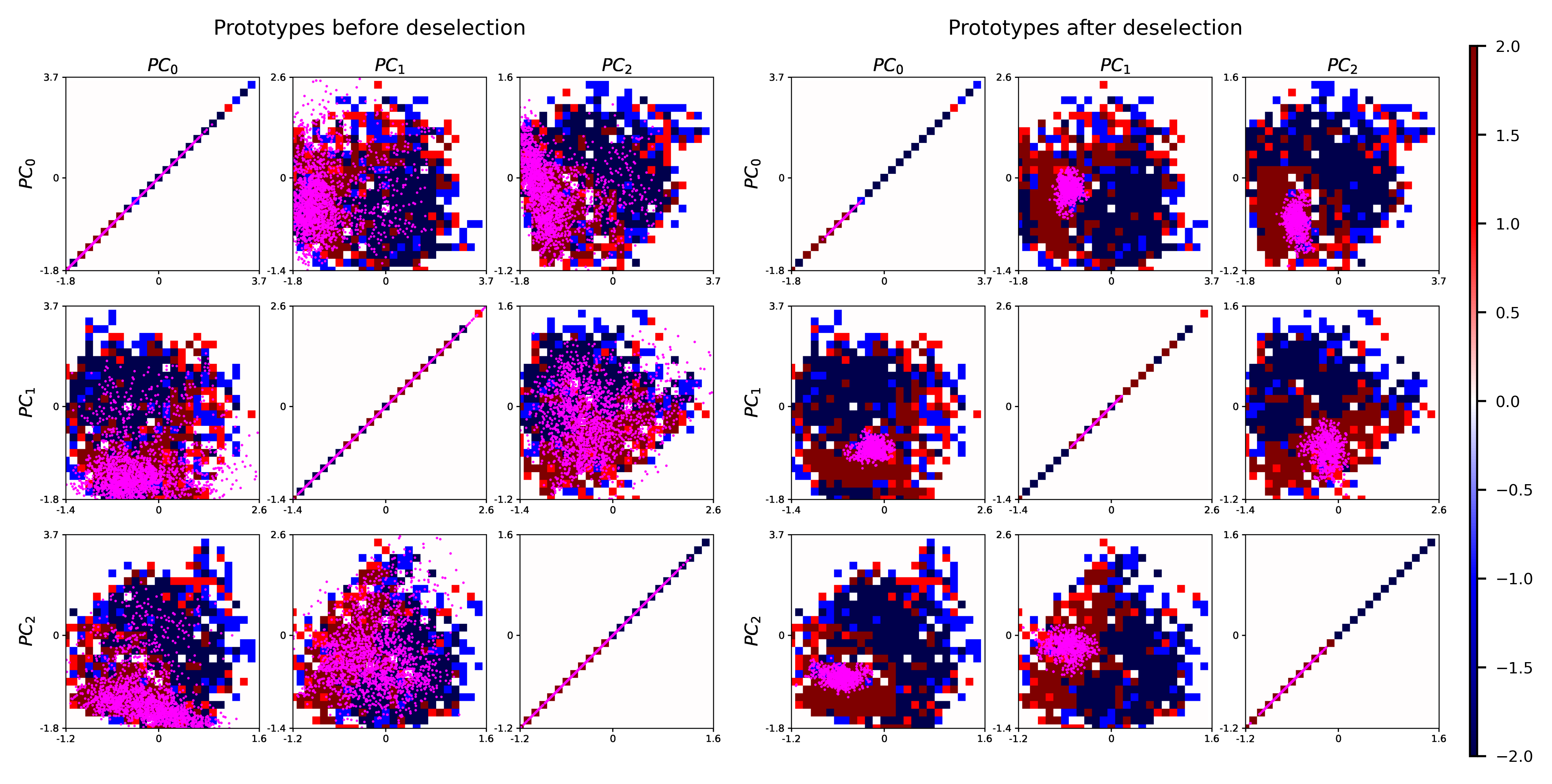}
\caption{Movement of prototypes upon iterative revision: The prototype vectors (magenta) in non-object areas (blue) move towards areas of object patches (red).}
\label{latent_effect}
\end{figure}

For condition two we apply our interactive procedure. It aims at moving prototypes in latent space to regions that encode object-related features. Inference adjustment is achieved by training the network with the presented deselection loss. After the first round between 38 and 77 prototypes revealed themselves as non-object prototypes. None of these prototypes covers the same image patch as before. The accuracy drops from .778 to an average of .732 and recovers to a level of .745 when the last layer is trained. The number of non-object prototypes converges to zero after five repetitions (\Figref{main_results}, lower left) while the accuracy increases: The average accuracy is then slightly higher with .739 before and .749 after fine-tuning of the last layer (\Figref{main_results}, lower right). The procedure yields the highest number of prototypes with near complete object-intersection as compared to prototype masking and the baseline condition (\Figref{main_results}, upper panel). After iterative revision 1583 of the 2000 prototypes show an overlap of at least 75\% with the object they represent: The prototypes become sensitive to object features (\Figref{main_results}, upper left).
\Figref{latent_effect} shows the effect on the distribution of prototypes. Each subplot shows the 2D subspace defined by the specified principal components of the grid vectors. The background indicates the relative density of object and non-object grid vectors in red and blue. The prototype vectors (magenta) move towards object areas.\\
 In the third condition we remove prototypes with no object-overlap by masking in the classification head (\Figref{network}). The average accuracy for the ten repetitions is $.778$ before revision and drops to $.77$ upon masking. After training of the last layer it mostly recovers. With an average of $.776$ it is marginally lower then before. No significant difference can be found using the T-statistic ($p=.29$). The accuracy is 2.7\% higher as compared to iterative revision where arguably a greater loss of information from non-object prototypes occurs as larger areas of feature space are excluded. However, iterative revision alters the overall distribution in favor of object-prototypes while masking removes only the prototypes that have no object-overlap at all (\Figref{main_results} upper right).\\

 The interactive procedure leads to a substantial increase in prototype-object intersection: The prototypes become sensitive to object features at the cost of only a marginal decrease in classification accuracy (below 3 \%).\\

\section{Conclusion}
\label{sec:conclusion}
 We show that the inference of prototypical part models cannot only be understood but also adjusted interactively solely by the interaction with concepts in latent space without the need for encoder training. Prototypes can be refined such that the prediction is restricted to meaningful features. The resulting prototypes are highly sensitive to object-features while the effect of spurious features is eliminated. Removing the effect of antitypes leads to a marginal loss in accuracy (1.4\% for masking and  2.9\% for revision). With the presented interactive procedure we ensure that object features are used only for the prediction. As compared to the  na\"{i}ve approach of masking the objects in the 6200 training images manually it requires considerably less effort (only 350 clicks). This number could potentially be further reduced if only a subset of prototypes was evaluated in each round. \\ As different strategies for inference exist future works should focus on scenarios that allow for a systematic quantitative comparison. Future research could also address configurations of prototypes, for example by interactive learning of deformable parts \parencite{branson_strong_2011}. Prototypes for the prediction of bounding boxes should be investigated. They could potentially help to model whole-part relationships in neural networks \parencite{hinton_how_2021}. Active learning with prototypes may also leverage potentials for increased labeling efficiency \parencite{gal_deep_2017}. Interactive prototype learning allows to alter the explanations of the model increasing trust in its predictions. As such it will arguably be a valuable tool for future applications of prototype networks.
\subsubsection*{Acknowledgements}
This research has received funding from the German Federal Ministry for Economic Affairs and Climate Action as part of the NaLamKI project under Grant 01MK21003D.
\printbibliography
\end{document}